\documentclass[11pt,a4paper]{article}
\usepackage{authblk}
\usepackage[hyperref]{naaclhlt2019}
\usepackage{times}
\usepackage{latexsym}
\usepackage{graphicx}
\usepackage{url}
\aclfinalcopy 

\title{Identifying Offensive Posts and Targeted Offense from Twitter}

\author{
Haimin Zhang,\textsuperscript{\rm 1}
Debanjan Mahata,\textsuperscript{\rm 1}
Simra Shahid,\textsuperscript{\rm 2}
Laiba Mehnaz,\textsuperscript{\rm 2} \\
Sarthak Anand,\textsuperscript{\rm 3}
Yaman Kumar,\textsuperscript{\rm 5}
Rajiv Ratn Shah,\textsuperscript{\rm 4}
Karan Uppal\textsuperscript{\rm 1} \\ \\
\textsuperscript{\rm 1}Bloomberg, USA,
\textsuperscript{\rm 2}DTU-Delhi, India,
\textsuperscript{\rm 3}NSIT-Delhi, India,
\textsuperscript{\rm 4}IIIT-Delhi, India,
\textsuperscript{\rm 5}Adobe, India\\
hzhang449@bloomberg.net, dmahata@bloomberg.net, simrashahid\_bt2k16@dtu.ac.in, \\ laibamehnaz@dtu.ac.in, sarthaka.ic@nsit.net.in, ykumar@adobe.com, \\rajivratn@iiitd.ac.in, kuppal8@bloomberg.net
}

\date{}

\begin{document}
\maketitle
\begin{abstract}
In this paper, we present our approach and the system description for Sub-task A and Sub Task B of SemEval 2019 Task 6: Identifying and Categorizing Offensive Language in Social Media. Sub-task A involves identifying if a given tweet is offensive or not, and Sub Task B involves detecting if an offensive tweet is targeted towards someone (group or an individual). Our models for Sub-task A is based on an ensemble of Convolutional Neural Network, Bidirectional LSTM with attention, and Bidirectional LSTM + Bidirectional GRU, whereas for Sub-task B, we rely on a set of heuristics derived from the training data and manual observation. We provide a detailed analysis of the results obtained using the trained models. Our team ranked 5th out of 103 participants in Sub-task A, achieving a macro F1 score of 0.807, and ranked 8th out of 75 participants in Sub Task B achieving a macro F1 of 0.695.
\end{abstract}

\section{Introduction}
\label{intro}
The unrestricted use of offensive language in social media is disgraceful for a progressive society as it promotes the spread of abuse, violence, hatred, and leads to other activities like trolling. Offensive text can be broadly classified as abusive and hate speech on the basis of the context and target of the offense. Hate speech  is an act of offending, insulting or threatening a person or a group of similar people on the basis of religion, race, caste, sexual orientation, gender or belongingness to a specific stereotyped community \cite{schmidt2017survey, fortuna2018survey}. Abusive speech categorically differs from hate speech because of its casual motive to hurt using general slurs composed of demeaning words. Both of them are the popular categories of offensive content, widespread in different social media channels.

With the democratization of the web, the usage of offensive language in online platforms is a clear indication of misuse of our right to `Freedom of Speech'. While censorship of free moving online content curtails the freedom of speech, but unregulated opprobrious tweets discourage free discussions in the virtual world making the problem of identifying and filtering out offensive content from social media an important problem to be solved for creating a better society, both in and out of the Internet.

Detecting offensive content from social media is a hard research problem due to variations in the way people express themselves in a linguistically diverse social setting of the web. A major challenge in monitoring online content produced on social media websites like Twitter, Facebook and Reddit is the humongous volume of data being generated at a fast pace from varying demographic, cultural, linguistic and religious communities. Apart from the problem of information overload, social media websites pose challenges for automated information mining tools and techniques due to their brevity, noisiness, idiosyncratic language, unusual structure and ambiguous representation of discourse. Information extraction tasks using state-of-the-art natural language processing techniques, often give poor results when applied in such settings \cite{ritter2011named}. Abundance of link farms, unwanted promotional posts, and nepotistic relationships between content creates additional challenges. Due to the lack of explicit links between content shared in these platforms it is also difficult to implement and get useful results from ranking algorithms popularly used for web pages \cite{mahata2015chirps}.

Interests from both academia and industry has led to the organization of related workshops such as TA-COS\footnote{\url{http://ta-cos.org/}}, Abusive Language Online\footnote{\url{https://sites.google.com/site/abusivelanguageworkshop2017/}}, and TRAC\footnote{\url{https://sites.google.com/view/trac1/home}}, along with shared 
tasks such as GermEval \cite{wiegand2018overview} and TRAC \cite{trac2018report}. The task 6 of SemEval 2019 \cite{offenseval} is one such recent effort containing short posts from tweets collected from the Twitter platform and annotated by human annotators with the objective of \textit{identifying expressions of offensive language}, \textit{categorization of offensive language} and \textit{identifying the target against whom the offensive language is being used}, leading to three sub tasks (A, B and C). We only participate in two of them for which we define the problems. 

\noindent \textit{\textbf{Problem Definition Sub-task A}} - \noindent \textit{Given a labeled dataset $D$ of tweets, the objective of the task is to learn a classification/prediction function that can predict a label $l$ for a given tweet $t$, where $l \in \{OFF, NOT\}$, OFF - denoting a tweet being offensive, and NOT - denoting a tweet being not offensive.}

\noindent \textit{\textbf{Problem Definition Sub Task B}} - \noindent \textit{Given a labeled dataset $D$ of tweets, the objective of the task is to learn a classification/prediction function that can predict a label $l$ for a given tweet $t$, where $l \in \{TIN, UNT\}$, TIN - denoting an offensive tweet targeted towards a group or an individual, and UNT - denoting a tweet that does not contain a targeted offense although it might use offensive language.}

Towards this objective we make the following contributions in this work:

\begin{itemize}
\item Train deep learning models of different architectures - Convolutional Neural Networks, Bidirectional LSTM with attention and Bidirectional LSTM + Bidirectional GRU, and report their results on the provided dataset. Our best model which ranked 5th in Sub-task A, is an ensemble of all the three deep learning architectures.
\item We perform an analysis of the dataset, point out certain discrepancies in annotation and show how undersampling directed by error analysis could be sometimes useful for increasing the performance of the trained models.
\end{itemize}

Next, we present previous works related to the task.

\section{Related Work}
Most of the previous works in this domain deals with the identification and analysis of the use of hate speech \cite{davidson2017automated}, and abusive languages in online platforms \cite{nobata2016abusive}. Abusive speech categorically differs from hate speech because of its casual motive to hurt using general slurs composed of demeaning words. A proposal of typology of abusive language sub-tasks is presented in \cite{waseem2017understanding}. Both abusive as well as hate speech are sub-categories of offensive language. Detailed surveys of the works related to hate speech could be found in \cite{schmidt2017survey} and \cite{fortuna2018survey}.

One of the earliest efforts in hate speech detection can be attributed to \cite{spertus1997smokey} who had presented a decision tree based text classifier for web pages with a  88.2 \% accuracy. Contemporary works on Yahoo news pages were done \cite{sood2012automatic}, and later taken up by \cite{yin2016ranking}. \cite{xiang2012detecting} detected offensive tweets using logistic regression over a tweet dataset with the help of a dictionary of 339 offensive words. Offensive text classification in online textual content have been tried previously for languages other than English, like German \cite{ross2017measuring}, Chinese \cite{su2017}, Slovene \cite{fiser2017}, Arabic \cite{mubarak2017abusive}, and in challenging cases of code-switched languages such as Hinglish \cite{mathur2018detecting}.  However, despite the various endeavors by language experts and online moderators, users continue to disguise their abuse through creative modifications that contribute to multidimensional linguistic variations \cite{clarke2017dimensions}.

\cite{badjatiya2017deep} used CNN based classifiers to classify hateful tweets as racist and sexist. \cite{park2017one} introduced a combination of CharCNN and WordCNN architectures for abusive text classification. \cite{gamback2017using} explored four CNN models trained on character n-grams, word vectors based on semantic information built using word2vec, randomly generated word vectors, and word vectors combined with character n-grams to develop a hate-speech text classification system. \cite{pitsilis2018detecting} used an ensemble of RNNs in order to identify hateful content in social media.

Some of the recent works in this domain has been on identifying profanity vs. hate speech \cite{malmasi2018challenges}, which highlights the challenges of distinguishing between profanity, and threatening language which may not actually contain profane language. On a similar direction there has been work on understanding the main intentions behind vulgar expressions in social media \cite{holgate2018swear}. Various approaches have been taken to tackle both textual as well as multimodal data from Twitter and social media in general, in order to build deep learning classifiers for similar tasks \cite{baghel2018kiki, kapoor2018mind, mahata2018detecting, mahata2018phramacovigilance, jangid2018aspect, meghawat2018multimodal, shah2017multimodal}.

\section{Dataset}
\label{data}
\begin{figure}[htbp]
\centering
\setlength{\belowcaptionskip}{-10pt}\includegraphics[width=0.45\textwidth]{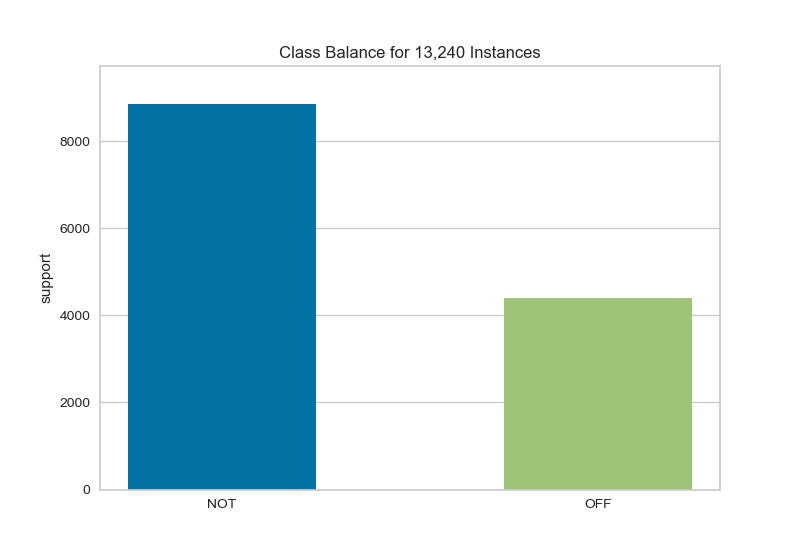}
\caption{Distribution of classes (OFF - Offensive and NOT - Not Offensive) for Sub-task A.)}
\label{fig:1}
\end{figure}

\begin{figure}[htbp]
\centering
\setlength{\belowcaptionskip}{-10pt}\includegraphics[width=0.45\textwidth]{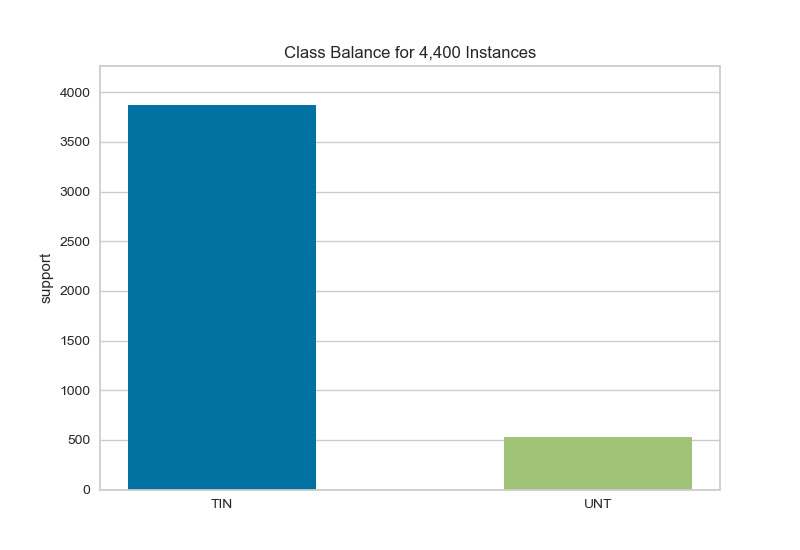}
\caption{Distribution of classes (TIN - Targeted Offense and UNT - Untargeted Offense) for Sub Task B.)}
\label{fig:2}
\end{figure}

The dataset provided for the tasks was collected through Twitter API by searching for tweets containing certain selected keyword patterns popular in offensive posts. Around 50\%  of the keyword patterns were political in nature such as `MAGA', `antifa', `conservative' and `liberal'. The other half were based on keyword patterns such as `he is', `she is', in combination with metadata provided by the Twitter API that marks a tweet to be `unsafe'. The annotation of the collected data was done using figure eight, which is a popular crowdsourcing platform. 14,100 tweets were selected in the final dataset with 13,240 provided as the training data and 860 as the test data. The details of the dataset, its collection process and annotation agreements could be found in \cite{OLID}. 

Figures \ref{fig:1} and \ref{fig:2}, shows the distribution of the classes in the subsets of the data provided for Sub-task A and Sub Task B, respectively. The distributions show the imbalance in class labels. We also took a detailed look at the dataset and found discrepancies between the definition of the classes as provided by the organizers and the actual annotations. The mislabeling was more prominent as an offensive post being labeled as not offensive. We observed such wrong annotations when performing manual error analysis on the predictions provided by an initially trained classifier, which was a simple Convolutional Neural Network. About 4 \% of the posts seemed to have been mislabeled, which we found through manual inspection and removed them from the training data. Here are few such examples. 
\begin{itemize}
\item  @user @user @user @user @user @user @user what a stupid incompetent devious and toxic pm ! may haven't you forgotten 17.4 million voters ? betray us at your peril ! you are eroding faith in democracy + destroying tory party ! you should go url. (\textbf{Original Label: NOT})

\item  angelina is so funny at rhe wrong times imngonna shoot this bitch uppdoals. (\textbf{Original Label: NOT})

\item  @user @user so and accusation by a libtarded trump hating liberal activist against a trump appointee doesnt make u wonder if the accusation was politically motivated in the slightest ? no ? this is why conservatives think u are all stupid . because u are . (\textbf{Original Label: NOT})	
\end{itemize}
This increased the performances of our trained models and could be considered as a heuristic based undersampling of the provided dataset.

\section{Experiments and Results}

We train different deep learning models for the Sub-task A and rely on heuristics learnt from the training data for Sub-task B. In this section we explain the steps taken for pre-processing data and training the predictive models and give a short description of the heuristics that we came up with after analyzing the data.

\subsection{Data Preprocessing}
\label{preprocess}
Before feeding the dataset to any machine learning model we took some steps to process the data. For all our experiments we used Keras\footnote{https://keras.io/} as the machine learning coding library. Some of the pre-processing steps that we took are:

\noindent \textbf{Tokenization} - Tokenization is a fundamental pre-processing step and could be one of the important factors influencing the performance of a machine learning model that deals with text. As tweets include wide variation in vocabulary and expressions such as user mentions and hashtags, the tokenization process could become a challenging task. We used the nltk's\footnote{https://www.nltk.org/api/nltk.tokenize.html} tweet tokenizer in order to tokenize the tweets provided in the dataset by overriding the default tokenizer provided in keras. 

\noindent \textbf{Cleaning and Normalization} - Normalization of tokens were also done using some hand-crafted rules. The \# symbol was removed from the tweets along with mapping few popular offensive words to a standard form. For example, `bi*ch', `b**ch', `bi**h', `biatch' were all mapped to `bitch', and `sob', `sobi*ch', were mapped to `son of bitch'. The @user tokens were removed. The hashtags that contained two or more words were segmented into their component words. For example \#fatbastard was converted to fat bastard. 
\subsection{Training Deep Learning Models}
In order to train deep learning models we need to provide the input as a matrix and the input words need to be mapped to their embeddings which provides richer semantic representation of words in comparison to the one-hot vectors. Each tweet is treated as a sequence of words and may vary in their lengths. We fix 200 as the max length and pad the input sequences in order to make their lengths fixed to 200. For, our experiments we used the 200 dimensional Glove embeddings\footnote{https://github.com/plasticityai/magnitude} trained on tweets and 400 dimensional Godin embeddings\footnote{https://fredericgodin.com/software/}. There was no significant difference in the results while training our initial models by using one over the other. Therefore for all our models as presented in this work we selected the Glove embeddings as the pre-trained word embedding of our choice due to its lower dimensions resulting in lesser training of weights in the neural network.

\begin{table*}[htbp]
\center
\setlength{\belowcaptionskip}{-10pt}\scalebox{0.80}{\begin{tabular}{|lll|}
\hline
\bf System & \bf F1 (macro) & \bf Accuracy \\ 
\hline
All NOT baseline & 0.4189 & 0.7209 \\
All OFF baseline & 0.2182 & 0.2790 \\
\hline
Convolutional Neural Network (on training data) & 0.8020 & 0.8387 \\
Bidirectional LSTM with Attention (on training data) & 0.7851 & 0.8246  \\
Bidirectional LSTM + Bidirectional GRU (on training data) & 0.7893 & 0.8301  \\
MIDAS Submission 1 on test data (CNN) & 0.7964 & 0.8395 \\
MIDAS Submission 2  on test data (Ensemble of CNN, BLSTM with Attention, BLSTM + BGRU) & 0.8066 & 0.8407 \\
\hline
\end{tabular}}
\caption{Results for Sub-task A.}
\label{tab:results-A-open}
\end{table*}

\begin{table}[htbp]
\center
\setlength{\belowcaptionskip}{-10pt}\begin{tabular}{|lll|}
\hline
\bf System & \bf F1 (macro) & \bf Accuracy \\ 
\hline
All TIN baseline & 0.4702 & 0.8875 \\
All UNT baseline & 0.1011 & 0.1125 \\
\hline
MIDAS Submission 1 & 0.6952 & 0.8667 \\
\hline
\end{tabular}
\caption{Results for Sub-task B.}
\label{tab:results-B-open}
\end{table}

We train the following architectures for Sub-task A having the parameters as explained next.

\noindent \textbf{Convolutional Neural Network} - Convolutional neural networks are effective in text classification tasks primarily because they are able to pick out salient features (e.g., tokens or sequences of tokens) in a way that is invariant to their position within the input sequence of words. In our model, we use three different filters with sizes 2, 3 and 4. For each filter size, 256 filters are used. A max pooling layer is then applied for each filter size. The resultant vectors are concatenated to form the vector that represents the whole tweet. A drop out layer with drop out rate 0.3 is applied before the input to the Multi Layer Perceptron with 256 neurons for classification. We also use a dropout layer after the embedding with dropout rate 0.3 to randomly drop words, which we find helpful to resolve overfitting issue. Sigmoid activation function is applied to the final layer.  

\noindent \textbf{Bidirectional LSTM with Attention} - Bidirectional LSTM (BLSTM) is an extension of LSTM in which two LSTM models are trained on the input sequence. The first on the input sequence as-is and the second on its reversed copy. This can provide additional context to the network and result in faster and sometimes better learning. They have shown very good results in sequence classification tasks. We use 64 LSTM units with 0.2 drop out, one attention layer is added on the sequence of result vectors from BLSTM. 128 neurons are used in the final Multi Layer Perceptron layer for classification. Sigmoid activation function is applied to the final layer.

\noindent \textbf{Bidirectional LSTM followed by Bidirectional GRU} - We use 64 LSTM units wrapped by a Bidirectional layer, 0.3 was the dropout rate, followed by a Bidirectional GRU with 64 GRU units also with 0.3 dropout. Then a max pooling and average pooling are used and concatenated before input to the final Multi Layer Perceptron layer with 128 neurons for classification. Sigmoid activation function is applied to the final layer.


For all three models we add a drop out layer after the embedding to randomly drop words, which we find helpful to address overfitting issue, and early stop is used with restoring the best model weights. Grid search is used to find the best parameters for each model. Table \ref{tab:results-A-open} presents the performance of each of these networks on the modified dataset as already explained in Section \ref{data}. 

Often, one solution to a complex problem does not fit to all scenarios. Thus, researchers use ensemble techniques to address such problems. Historically, ensemble learning has proved to be very effective in most of the machine learning tasks including the famous winning solution of the Netflix Prize. Ensemble models can offer diversity over model architectures, training data splits or random initialization of the same model or model architectures. Multiple average or low performing learners are combined to produce a robust and high performing learning model. We do the same in our experiments. We combine the trained deep learning models having different architectures as an ensemble by averaging their final predictions. We had also tried the stacked ensemble approach as explained in \cite{mahata2018phramacovigilance}. But it didn't give promising results in first few iterations. Moreover, it was computationally expensive and due to lack of sufficient time we, did not go further in that route.

\begin{figure}
\centering
\setlength{\belowcaptionskip}{-10pt}\includegraphics[width=0.4\textwidth]{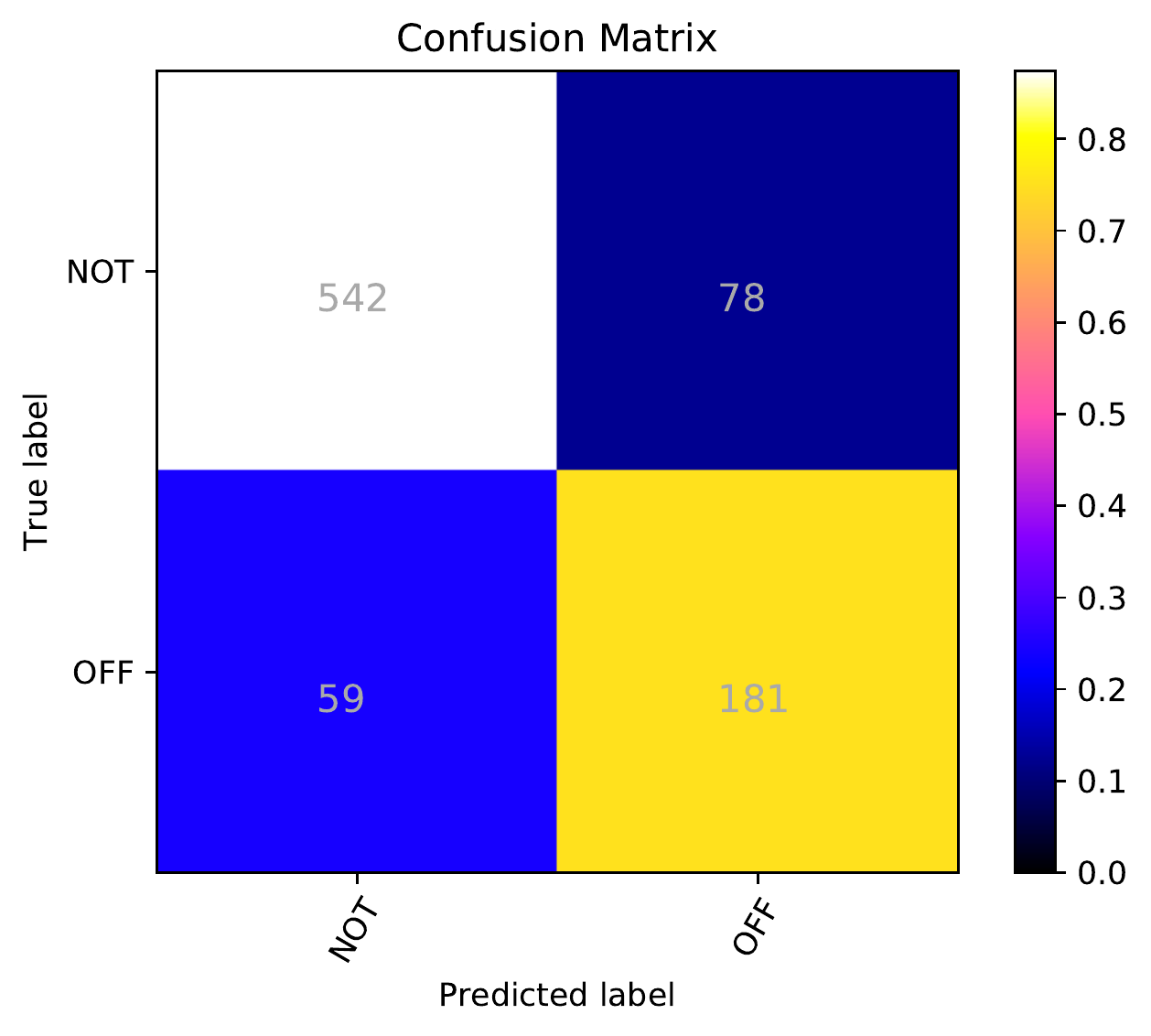}
\caption{Confusion Matrix for MIDAS submission 2 for Sub-task A.}
\label{fig:3}
\end{figure}

Our ensemble model performed better than the individual models and was also submitted to the competition, which was finally ranked 5th amongst 103 participants. Figure \ref{fig:3} presents the confusion matrix of our submission for Sub-task A. Some of the samples from the training dataset, which were very hard for our final model to predict are:

\begin{itemize}
\item More like \#Putin every day. \#MAGA URL (OFF)
\item @USER Hitler would be so proud of David Hogg trying to disarm American citizen so when Democrats come to power-we are helpless And cannot defend ourselves-; that's why we have they AR15's (NOT)
\item @USER good job (sarcasm). Also great they have gun control laws it’s saving lives! (More sarcasm). (OFF)
\end{itemize}

\begin{figure}
\centering
\setlength{\belowcaptionskip}{-10pt}\includegraphics[width=0.4\textwidth]{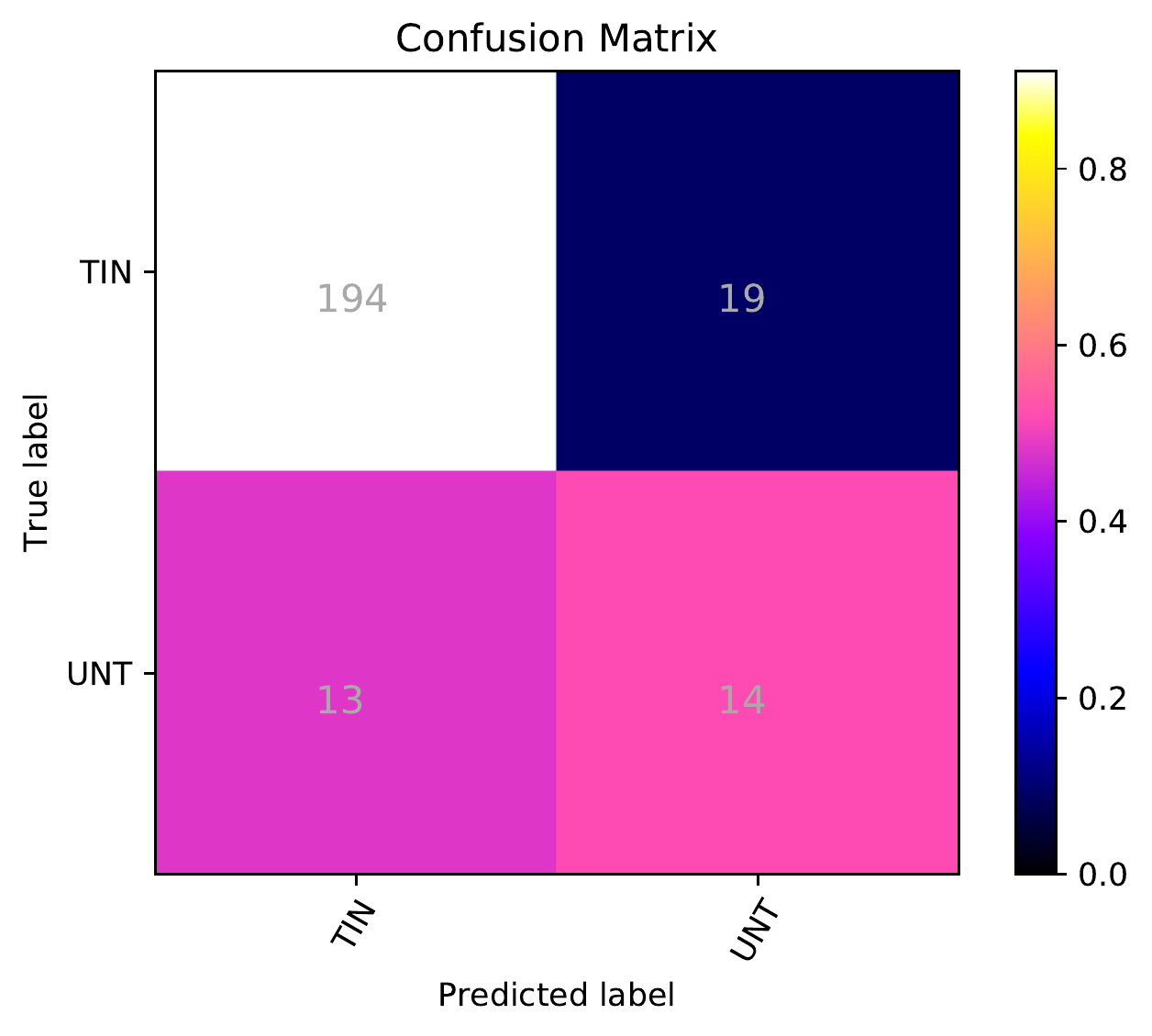}
\caption{Confusion Matrix for MIDAS submission 1 for Sub-task B.}
\label{fig:4}
\end{figure}

\subsection{Heuristics for Sub-task B}
Due to lack of time from our part, we were not able to train good machine learning models for Sub-task B. The preliminary models that we trained showed performances that was similar to that of a random model biased by the class distribution of the training data. The training dataset for Sub-task B was highly imbalanced which was a major challenge. We would like to have an in depth look at Sub-task B in the near future. 

For the sake of submission to the competition we came up with certain heuristics in order to decide whether an offensive post is targeted or not. We skipped the pre-processing part of the tweets that we did before training the machine learning models as described in  Section \ref{preprocess}. We looked at the frequency distribution of words and hashtags in the training dataset as well as observed the patterns of the posts. After doing that we did find that some of the hashtags like `\#maga', `\#liberals', `\#kavanaugh', `\#qanon', etc were frequently occurring. and so are some of the tokens like `antifa', `president', `trump', `potus', `liberals', `conservatives', `democrat', `nigga', `gay', `jew'. Top 100 such tokens and hashtags were compiled after eliminating some of them manually if they didn't make any sense, for example some unwanted stop words. We also extracted POS tags of the tweets using TweeboParser\footnote{http://www.cs.cmu.edu/~ark/TweetNLP/} and extracted named entities (only PERSON, ORG, LOCATION, FACILITY) using SpaCy\footnote{https://spacy.io}. We framed our final heuristic based on the following rules:

\begin{itemize}

\item If the post includes any of the 100 hashtags then it is considered as targeted offense (TIN).
\item else if the post includes any of the 100 tokens then it is considered as targeted offense (TIN).
\item else if no named entity in the post and no Personal Pronoun and Proper Nouns are present in the post then it is a untargeted offense (UNT).
\item else if the post has he/she is, you are, he she then it is considered as  targeted offense (TIN).
\item else if the post has pattern ' Starts with hashtag followed by verbs and named entity' then it is considered as targeted offense (TIN).
\item else If there is a named entity then it is considered as targeted offense (TIN).
\item all other cases are considered as untargeted offense (UNT).
\end{itemize}

We do not think this to be a robust model and it was only possible to come up with the heuristics because there were certain patterns in the dataset that was very obvious to bare human eye. Given that the dataset is very small, these heuristics can never scale well. One of the reasons behind discovering such patterns could also be because of the way the dataset was collected. Now that we know how it was collected as explained in Section \ref{data}, these patterns make more sense and it does explain why we could perform reasonably well even though we came up with such naive patterns in haste. Figure \ref{fig:4} presents the confusion matrix of our submission for Sub-task B and Table \ref{tab:results-B-open} presents the performance on the test dataset.

\section{Conclusion and Future Work}

In this work, we report our models and their respective performances in Sub-task A and B of SemEval-2019 Task 6 OffensEval: Identifying and Categorizing Offensive Language in Social Media. We showed how an ensemble of deep learning models performed well in the provided dataset and was ranked 5th in the competition in Sub-task A. Due to the inherent biases in collecting the dataset we believe that we were able to come up with naive heuristics for Sub-task B and was able to rank 8th in the competition. 

In the future we would like to solve Sub-task B using a machine learning approach. We would also like to look at other machine learning architectures and ensemble methods for the different sub tasks in the competition. Out of three sub tasks, we were able to attempt only two of them. In the near future we would like to tackle the problem posed in Sub-task C. Some of the other areas that could be explored are cleaning the dataset by correcting the annotations and studying the problem of inherent biases that can occur in samples collected based on keyword patterns.

\bibliography{semeval}
\bibliographystyle{acl_natbib}

\end{document}